\documentclass[conference]{IEEEtran}
\IEEEoverridecommandlockouts
\usepackage{cite}
\usepackage{amsmath,amssymb,amsfonts}
\usepackage{algorithmic}
\usepackage[ruled,vlined]{algorithm2e}
\usepackage{graphicx}
\usepackage{textcomp}
\usepackage{xcolor}
\def\BibTeX{{\rm B\kern-.05em{\sc i\kern-.025em b}\kern-.08em
    T\kern-.1667em\lower.7ex\hbox{E}\kern-.125emX}}
\begin{document}

\title{Human-in-the-loop online multi-agent approach to increase trustworthiness in ML models through trust scores and data augmentation\\
}

\author{Gusseppe Bravo-Rocca\IEEEauthorrefmark{1}\IEEEauthorrefmark{2},
\IEEEauthorblockN{Peini Liu\IEEEauthorrefmark{1}\IEEEauthorrefmark{2},
Jordi Guitart\IEEEauthorrefmark{1}\IEEEauthorrefmark{2},
\\
Ajay Dholakia\IEEEauthorrefmark{3},
David Ellison\IEEEauthorrefmark{3},
Miroslav Hodak \IEEEauthorrefmark{3}
}
\IEEEauthorblockA{\IEEEauthorrefmark{1}Barcelona Supercomputing Center, Barcelona, Spain}
\IEEEauthorblockA{\IEEEauthorrefmark{2}Universitat Polit\`ecnica de Catalunya, Barcelona, Spain}
\IEEEauthorblockA{\IEEEauthorrefmark{3}Lenovo Infrastructure Solutions Group, Lenovo, Morrisville, NC, USA}

E-mail: \{gusseppe.bravo,peini.liu,jordi.guitart\}@bsc.es,\{adholakia,dellison,mhodak\}@lenovo.com}
\maketitle

\begin{abstract}
Increasing a ML model accuracy is not enough, we must also increase its trustworthiness. This is an important step for building resilient AI systems for safety-critical applications such as automotive, finance, and healthcare. For that purpose, we propose a multi-agent system that combines both machine and human agents. 
In this system, a checker agent calculates a trust score of each instance (which penalizes overconfidence and overcautiousness in predictions) using an agreement-based method and ranks it; then an improver agent filters the anomalous instances based on a human rule-based procedure (which is considered safe), gets the human labels, applies geometric data augmentation, and retrains with the augmented data using transfer learning. We evaluate the system on corrupted versions of the MNIST and FashionMNIST datasets.
We get an improvement in accuracy and trust score with just few additional labels compared to a baseline approach.
\end{abstract}

\begin{IEEEkeywords}
Trust, multi-agents, robust AI, data centric
\end{IEEEkeywords}

\section{Introduction}
In Machine Learning (ML), the main focus has been given to improving the accuracy of ML models, driven by benchmarks that aim to accelerate the development of new models. However, this is not enough to have reliable models ready to be used in real complex situations. In particular, it is not clear whether one should really trust the predictions of a model in applications where every decision carries some responsibility \cite{ai_clinics}. For example, a model that assists a doctor in a medical diagnosis based on scanned images can reduce diagnosis times to serve more patients and also reduce human mistakes. However, what if the model provides a wrong diagnosis? Worse yet, what if the wrong prediction delivered has a high percentage of confidence? Clearly, the model trustworthiness is a critical point to be analyzed and taken into account. Achieving trustworthy AI requires addressing many issues such as accountability, fairness, privacy, and safety \cite{trustworthy1}. 


To improve trust in ML models, many methods opt to first improve the accuracy of the model, by modifying internally its structure, and consequently improve the trust. While it is true that an improvement in accuracy helps to improve trust, it does not guarantee it (i.e., unbalanced datasets, fraud detection) \cite{unbalanced}. A better way is to interact with the model externally in such a way that, through continuous monitoring and evaluation, small gains in trust can be achieved with less human effort. 

\begin{figure}
        \centering
    \includegraphics[width=0.95\linewidth]{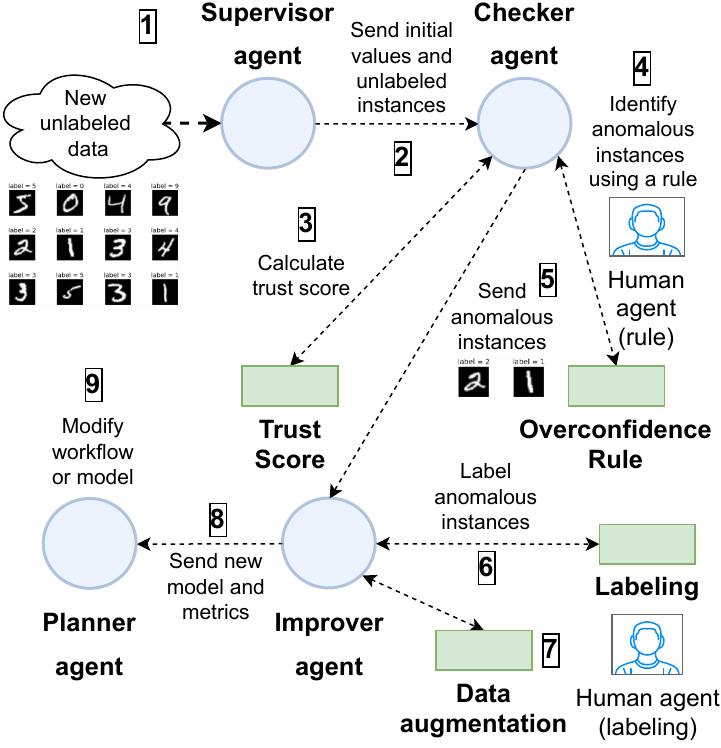}
    \caption{Overview of the proposed multi-agent system.}
    \label{fig:scenario1} 
\end{figure}

In this paper, we propose a multi-agent system that combines both machine and human agents to improve ML models trustworthiness. As shown in Fig. \ref{fig:scenario1}, a set of agents works together to quantify the trust on the set of predicted samples, in such a way that, given new sets of corrupted data to be inferred, they can detect untrustworthy samples and improve the model (and the training data) over time. The Supervisor agent is in charge of obtaining the initial values of the system, e.g., the current model, the training set, the training code, etc. With these values, the Checker agent trains a model that calculates a trust score for each individual instance, and applies a rule to identify anomalous instances. These anomalous instances are received by the Improver agent, which is assisted by a human agent to label them, applies data augmentation, and retrains the model using transfer learning from the weights of the previous model and the new training set (adding augmented labeled samples). This process is performed iteratively during the inference stage, so that the model can face situations similar to those it would encounter in real life. 



The remainder of this paper is as follows: Section \ref{sec:background} states the background and related work regarding the role of humans in ML, joint work of human and machine agents, agents in production systems, and trust in ML. Section \ref{sec:multiagent} declares the features and formal definition of the multi-agent system including its architecture. In Section \ref{sec:experiments-results}, we discuss the evaluation results on two datasets to demonstrate the system usefulness when using agents. Finally, the conclusions and new ideas for future work are presented in Section \ref{sec:conlusions-future-work}. 

\section{Background and Related work}
\label{sec:background} 

\subsection{The role of humans in learning systems}
The role of end-users in learning systems is often relegated to their usage, excluding the participation in their continuous improvement. Typically, the end-user's job is limited to supplying data, solving domain questions, or giving feedback on the learnt model. A more interesting alternative arises when the learning cycle is faster (user's feedback is quickly considered in the model), more focused (not the whole model is updated, only a particular feature), and with incremental model updates (changes are small, the model does not change drastically with each update) than the traditional ML process \cite{Amershi_2014}. This allows users with little expertise to explore and adapt the behavior of the model interactively, drastically reducing the intervention of experts (e.g., data scientists). Examples of such users' interaction with learning models are given in product recommendations in Amazon, movie recommendations in Netflix, and song recommendations in music services like Spotify, where the system asks its users about their preference on certain aspects of the product, the user responds with feedback (e.g., five stars punctuation, click on the "like" button, bad movie review, etc.), and the system quickly incorporates this feedback to improve the recommendation model. 

This idea of including the human in the learning process has been used in some works \cite{caruana_meta,human_interv}, including image segmentation \cite{inter_ml}, where the user corrects the segmentation made by the system by painting pixels as foreground or background in an image. This quick intervention allows the model to correct its prediction by creating new data in the most problematic areas. Our work takes this idea and extrapolates the model-human interaction to an agent-human one.

\subsection{Thinking fast and slow for machine and human agents}
\begin{figure}[htbp]
\centerline{\includegraphics[width=1.0\linewidth]{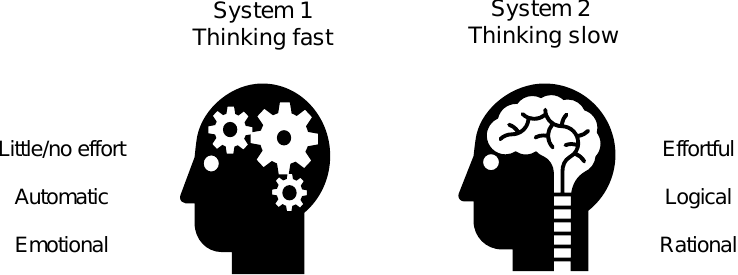}}
\caption{System 1 (e.g., answer to 2+2) vs. System 2 (e.g., answer to 32x231).}
\label{fig:system1y2}
\end{figure}

Kahneman’s research on human behavior \cite{kahneman2011thinking}, specifically, on reasoning and decision-making, describes two ways in which the brain forms thoughts, namely \textit{System 1} and \textit{System 2}, which are shown in Fig. \ref{fig:system1y2}. \textit{System 1} operates intuitively, unconsciously, automatically, with no effort, and is fast. For instance, it comes up when performing the operation 2+2 or recognizing a friend's face; all of them require little or no effort. \textit{System 2} involves concentration, attention, conscious reasoning, and is slow. For instance, it occurs when performing the operation 245x179 or giving someone your email. \textit{System 1} is mainly guided by intuition and \textit{System 2} by deliberation. 

These ideas can be applied to AI as well. \textit{System 1} mimics how ML works and \textit{System 2} mimics symbolic AI. That is to say, \textit{System 1} is data-driven and \textit{System 2} is knowledge-driven. Similar to ML, \textit{System 1} is good at sensing and reading, i.e., perceptual tasks. Conversely, \textit{System 2} is grounded on planning, logic, search, and explicit knowledge \cite{booch2020thinking}. 

In our work, we use these ideas to develop a hybrid approach to improve a model over time using learning (fast) agents and symbolic (slow) agents. This approach is applied on both offline (just after training) and online (production) phases. For instance, a learning agent can be a drift anomaly detector, which detects out-of-distribution samples in the new data (inference stage). From these anomalous instances, the system can learn some insights (e.g., meaningful instances) about how to update the model. This learning process is fast (no human intervention required). A symbolic agent can take advantage of these insights and, for instance, give these instances to a human to create labels so that retraining the model leads to better accuracy. The symbolic agents are meant to be more logical, which allows humans to intervene in workflow improvement (enabling a human-in-the-loop approach). Of course, this is a slow step that requires some effort but it is needed to create trustworthy AI systems \cite{brundage2020trustworthy}.

Other related works exploit those hybrid approaches. An example is the Neuro-Symbolic Concept Learner \cite{mao2019neurosymbolic} which learns images and words by reading paired questions and answers that are then translated to symbolic programs. This combination allows the model to generalize better with only few examples. We share this same idea of having a compound system, however, we consider agents that interact externally without modifying internally the model's architecture. 





\subsection{Agents in production systems}
The agent technology has been used for a while for a myriad of applications such as robotics \cite{veloso99}, automated online assistants \cite{iJET142}, and improving coordination of human groups \cite{2017Natur.545..370S}, among others. An agent is an entity that lives in an environment and has its own goals, knowledge, and actions \cite{russel2010}. This agent models itself, its interactions, and the environment. If other agents appear, they are considered part of that environment \cite{MASsurvey}. In contrast, a multi-agent system models other agents' goals and actions directly via communication or by using the environment. In our work, we use the structure of a multi-agent system to model a direct communication between different agents (see Section \ref{sec:multiagent}).

Khalil et al. \cite{mas-learning} have explored the integration of ML and multi-agent systems through information sharing and Q-Learning to learn the behaviors. However, to maximize the reward, a large space of scenarios has to be explored, requiring a lot of data and computation. Our approach is to have pre-established behaviors to reduce the overload of states. To implement such a structure, we take advantage of microservices in contrast to the bunch of tools and languages that were used for such a goal \cite{Bellifemine1999JadeA,agentspeak,agentfactory}. 

Microservices allow building systems at the scale that is needed for creating a multi-agent system. Basically, they provide an isolated state, easy distribution, elasticity to best fit a scenario, automated management, and loose coupling (weak associations) \cite{microservices}. These features share many commonalities with the characteristics of agents \cite{agentmicroservice}, 
 and allow agents to have a live supervision over the workflow because they can perform actions by themselves all the time or when required.


Particularly, we use this 'prior' behavior to detect a new inference request (Supervisor agent), identify untrustworthy examples on this request (Checker agent), perform data augmentation and retraining (Improver agent), and change the production model (Planner agent) by a newer one (the one retrained using the new augmented and labeled instances).

\subsection{Measuring trust in ML models}

Trust in ML is an important ingredient to allow learning systems to go beyond simple predictions, that is, make these systems work closer with people on critical applications. To evaluate the trustworthiness of a model, similarly as done with the accuracy, we need a scalar value to compare with. For that purpose, we use the \textit{NetTrustScore} \cite{wong2021really}.

That calculation needs some prerequisites, namely, \textit{question-answer trust}, \textit{trust density}, and \textit{trust spectrum}. Equation (\ref{eq:question-answer}) quantifies the \textit{question-answer trust}, e.g., the trust when predicting whether a picture shows a cat, by calculating the trustworthiness of an individual instance $x$ (the question) for one label $y$ (answer) based on its behavior under correct and incorrect labels $z$ (oracle answers), that is, where $y$ matches $z$ ($R_{y=z \mid M}$) or where does not ($R_{y \neq z \mid M}$) over the space of all possible questions $x$ given a model $M$. Equation (\ref{eq:density}) measures the \textit{trust density}, e.g., the distribution of trust when predicting many pictures of a cat, by calculating the density distribution of question-answer trust over a set of instances (e.g., training set), in other words, the trust behavior of a model $M$ for a specific answer scenario $y$. Finally, Equation (\ref{eq:spectrum}) measures the \textit{trust spectrum}, e.g., the average overall trust for class 'cat', by calculating the average question-answer trust across all the $N$ instances for a given class $z$.

\begin{equation}
Q_{z}(x, y)= \begin{cases}C(y \mid x)^{\alpha} & \text { if } x \in R_{y=z \mid M} \\ (1-C(y \mid x))^{\beta} & \text { if } x \in R_{y \neq z \mid M}\end{cases}
\label{eq:question-answer} 
\end{equation}

\begin{equation}
F(Q_{z}) = \{q \in Q_{z}(x, y), \forall x \in X, z: answer\}
\label{eq:density} 
\end{equation}

\begin{equation}
T_{M}(z)=\frac{1}{N} \sum_{x=1}^{N} Q_{z}(x)
\label{eq:spectrum} 
\end{equation}


Given those, the \textit{NetTrustScore}, which measures the trust spectrum over all the $L$ classes (e.g., the overall trust for classes 'cat', 'dog', ...), is formalized as follows:

\begin{equation}
\textit{NetTrustScore}=\frac{1}{L} \sum_{z=1}^{L} T_{M}(z)
\label{eq:NetTrustScore} 
\end{equation}

We interpret the \textit{NetTrustScore} as the average scalar value that summarizes the overall trustworthiness of a ML model. It indicates how well the model's confidence behaves under all the possible answer (label, class) scenarios that might occur. On our evaluation, we use this metric to verify if a model improves its trustworthiness on a fixed group of instances.

Other related works explore how to obtain precise probability estimates for classification problems using calibrated binary probability estimates \cite{score_to_prob}. That is, based on two-class probability estimates, apply this measure to any classifier that ranks outputs by a number (which can be interpreted as a kind of trust). However, this method is used mainly for classical ML models, like Naive Bayes and Support Vector Machine, and has not been yet explored on Neural Networks. Similarly, classical target learners were applied in \cite{predict_prob}. Other works have targeted modern deep networks, like ensembles of neural networks \cite{deep_ensembles, uncertainty_esti}, and proposed to use an Average Early Stopping (AES) algorithm that tries to take advantage of the confidence probabilities as an ordinal ranking. However, these approaches are computationally expensive and yield lower interpretation. We choose \textit{NetTrustScore} because of its simplicity (simple equations), interpretability (only two concepts overconfidence and overcautiousness), and celerity (only the confidence probability is needed).

\section{Multi-agent system for trustworthy models}
\label{sec:multiagent}
This section describes our approach to increase trustworthiness in ML models as a multi-agent microservice system that combines learning and symbolic agents. 

\subsection{Agent properties}
In our current design, each agent is reactive with some limited internal state that is saved in a tracking repository. This means that the agents lack the deliberation to devise a long-term plan by themselves; instead, they keep an ongoing interaction with the environment (new input data and other agents) to act upon any change in time and respond timely. Additionally, each agent is rational because it is programmed to always perform the correct actions, the most appropriate to reach its objective, as a response to the information perceived by its sensors (new data and messages from other agents). 



Some agents have a learning behavior, for instance, the Checker, whose mechanism of detecting anomalies (whose trust score is considered anomalous) requires learning historical information about training data. Other agents, such as the Improver, have a symbolic behavior in the form of heuristics (rules that allow establishing the thresholds of the model trust) and in the use of human labeling to pick the instances with high uncertainty (anomalous cases of overconfidence).

Moreover, the agents have a temporal continuity, that is, the agents are available 24/7 to answer requests. However, the most important property is the agents' social ability to interact with others via some kind of agent-communication protocol. This ability builds on some coordination mechanism that our system must have, and that fulfills the following properties.

First, the agents are cooperative and respect the benevolence assumption. All the agents share a common goal and there is no conflict between them.
This implies that all that matters is the overall goal and not the individual one. This simplifies the optimization of the system utility as there is only one global utility and not one for each agent. For instance, in our system, the global goal is to improve the model trust. The benevolence property allows the designer to simplify the task as there is not conflict in the behavior nor in the communication.

Second, the agents have explicit communication and have the intention to exchange messages. In our system, they only react to the performative messages, such as QUERY, INFORM, PROPOSE, etc. A completely reactive system would react to each agent's behavior directly (implicit communication), and would require some deliberative reasoning to understand those conducts, which is not our case. For instance, the communication between the Supervisor agent and the Checker agent is explicit as the latter agent reacts once a certain type of performative message is received. 

Finally, the agent's interaction is driven by certain meta-negotiation. Our system implements some limited features from the Contract Net Protocol (CNP) \cite{contractnet}. In the original CNP, an agent acts as a manager for a limited time and makes a task announcement. If there is no knowledge about the capabilities of each agent, a general broadcast is performed to all the agents. Then, each receiver agent evaluates its own specialized knowledge and, if a capability is found, it submits a bid (offer). Once the manager selects the appropriate agent based on the information in the bids, it submits an award message and the selected agent becomes a contractor for that task. However, in our system, we already have some assumptions about each agent's beliefs. For example, as the Supervisor agent knows which agents should do the checking (Checker agents), it can issue a limited broadcast to only these candidates. Even more efficient is that the Supervisor agent performs a directed contract, because neither an announcement is needed nor bids are submitted. As a result, our system sends simple messages, reducing response times. 


\subsection{Agent definition}
To give a formal definition of the agents, we assume that the training data increases or improves in quality, sequentially and in discrete time, as the model receives requests. Additionally, the agent's actions, i.e., its conclusions, influence the future performance of the model. Given those premises, let $s^{i}_{t}$ denote the state of agent $i \in \mathcal{A}$ (set of agents) at time $t$, $a^{i}_{t}$ denote the action taken by agent $i$ at time $t$, $F^{i}_{t}$ denote the external flow of information (represented by messages $e$ that refer to requests $\tau$) sent by other agents that affects the behavior of agent $i$ at time $t$ (in particular, its decision about $a^{i}_{t}$) ($F^{i}_{t}=(e^{m, i}_{t}(\tau^{m})),  \forall m \in \mathcal{A}\setminus \{i\}$), $s^{i}_{t+1}$ denote the next state at time $t+1$ and $r^{i}_{t}$ the reward (in our case, the net-trust score) at time $t$, both obtained when agent $i$ applies $a^{i}_{t}$ using the information provided by $s^{i}_{t}$ and $F^{i}_{t}$ at each time $t$ (($(s^{i}_{t}, a^{i}_{t}, F^{i}_{t}) \implies s^{i}_{t+1}$), (($s^{i}_{t}, a^{i}_{t}, F^{i}_{t}) \implies r^{i}_{t}$)). 
Then, the goal is to find a sequence of actions $\left\{a^{i}_{t}\right\}$ to maximize the cumulative reward $\sum_{t=0}^{T} r^{i}_{t}$ given an initial state $s^{i}_{0}$. For instance, $a^{i}_{t}$ can be updating the model, training a Checker to detect distribution drift, or asking for human labels at the right time.

\subsection{Architecture of the multi-agent system}
The internal behavior of the agents in this paper builds on a previous work \cite{scanflow} that considers model debugging through static nodes in a graph. In a nutshell, we propose to convert those static nodes into dynamic agents (the Checker and Improver) and add two agents with new responsibilities (the Supervisor and Planner). A detailed description of each of these agents is provided in the following sections.

\begin{figure*}
        \centering
    \includegraphics[width=0.8\linewidth]{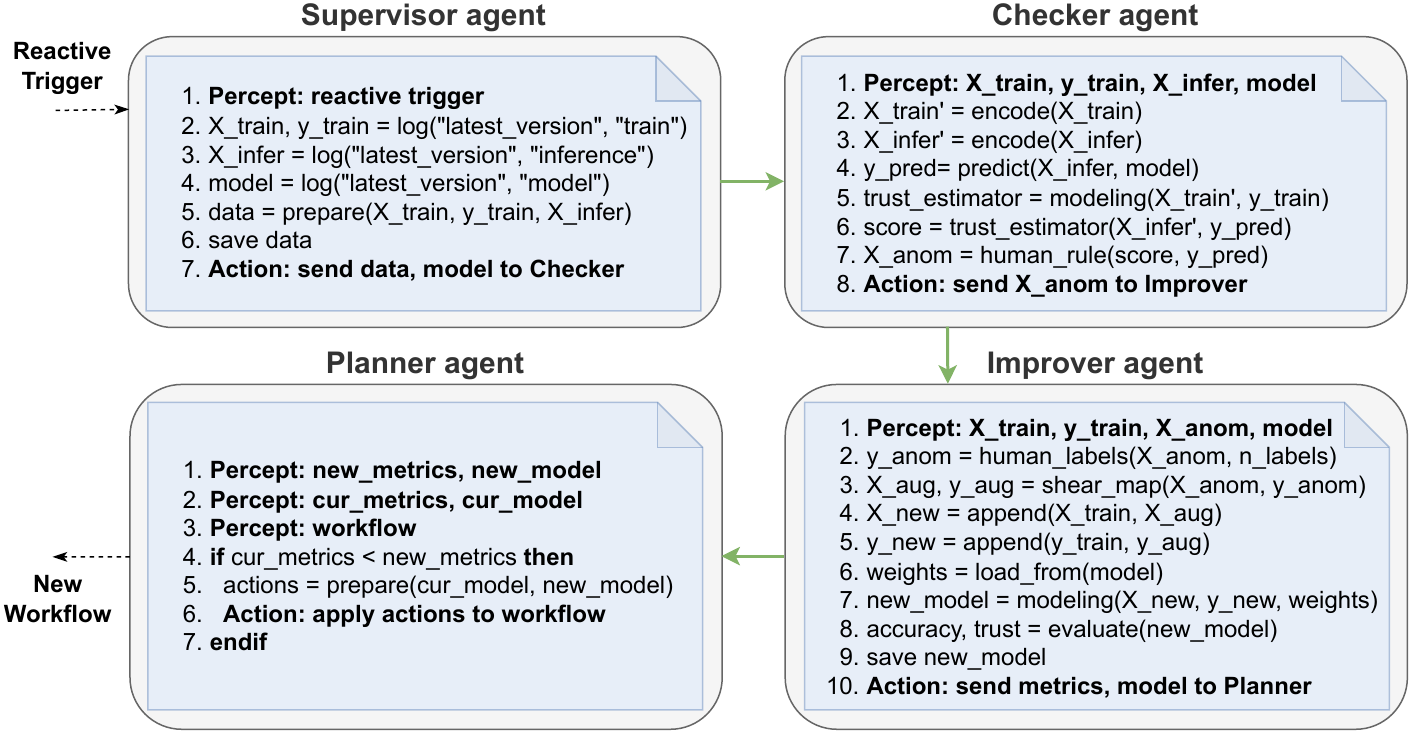}
    \caption{High level multi-agent execution. Figure shows the interaction between Supervisor, Checker, Improver, and Planner agents to enhance the model trust.}
    \label{fig:scanflow-extended-execution} 
\end{figure*}\hfill


\subsubsection{Supervisor (Initializing and starting the MAS)}
As shown in Fig. \ref{fig:scanflow-extended-execution}, this agent is in charge of getting the initial metadata, such as latest model, training data, etc., which are needed to proceed further in the debugging process. It starts the multi-agent system by sensing a stimulus (for instance, scheduling time or amount of inference samples) and exposes the metadata to the Checker, which will perform the detection and then send its outcome to the Improver, which eventually will send feedback to the Planner to perform the required actions to modify the model in production. 



  

  
  


\subsubsection{Checker (Untrustworthy points and rule-based threshold)}
As shown in Fig. \ref{fig:scanflow-extended-execution}, the Checker agent performs three steps, namely reducing the dimensionality of the input data using an autoencoder (the trust score estimator works best for low to medium feature space size), modeling a trust estimator to measure the agreement in the predictions (in the form of a trust score) between the model and a modified version of a nearest-neighbor (k-NN) classifier \cite{jiang2018trust}, and applying a rule-based procedure to assess what ranges of trust score values and prediction probabilities are considered not trustworthy.

As the labels are unknown in the inference stage, the Checker agent models a complementary unsupervised trust score estimator. Let be $h$ the model, $(X, Y) = (x_1, y_1), ..., (x_n, y_n)$ the training data, $x'$ the inference sample, and $K$ a k-NN estimator trained on $(X, Y)$ that estimates a density set (k-nearest neighbors of a class without outliers), the trust score of $x'$ is defined as shown in (\ref{eq:trust_score}), that is, it is the ratio between the distance from the inference sample to the density set of the nearest class different from the predicted class ($\hat{K}$), and the distance from the inference sample to the density set of the class predicted by $h$. 

\begin{equation}
T(h, x') = d(x', \hat{K}(h(x'))) / d(x', K(h(x')))
\label{eq:trust_score} 
\end{equation}

This trust score and the confidence of the predictions are then used to rank them according to a rule-based procedure (provided by a human agent) that tries to identify anomalous predictions that are considered not safe. If the model yields a prediction with high confidence (e.g., probability between 0.65 and 0.95) but with a low trust score (e.g., lower than 1), then this behavior is considered as overconfident, and the model should not be trusted despite its high confidence. The probabilities above 0.95 are considered trustworthy predictions. It is worth remarking that the Checker uses this trust score because it does not need test labels and is meant only from individual instances rather than the whole set of data (in contrast to NetTrustScore previously defined). 

  
 
 
  
  


\subsubsection{Improver (Human labeling and data augmentation)}

A form of human intervention that has worked very well in supervised learning problems is labeling. This technique allows business experts or product owners to label data to create datasets ready for the creation of ML models. Similarly, data augmentation came up as a way to artificially augment data in cases where datasets are small. ML models are susceptible to the quality of labeling and the number of samples. Therefore, applying the correct, unambiguous, and consistent labeling \cite{label_amb} is as important as augmenting data that follows the same training distribution, without adding noise \cite{importance_dataaug}.

The difficulty of applying these two techniques depends on the type and size of the data, i.e., structured or unstructured, small or big. In the case of unstructured data (images, audio transcripts, videos, etc.), people can label more data with less effort \cite{easy_label_human} and data augmentation is easier to perform (e.g., rotate or apply contrast to an image). On the other hand, when using structured data (tabular data), augmentation is more difficult, since each of the characteristics of an instance must be considered. Also, labeling structured data is more tedious in the sense that the person has to consider many variables to decide on a class (it is easier to look at an image and decide than to look at all the variables of a table). Regarding the data size, when you have small datasets, each label is critical to the model, i.e., small changes have a greater importance in the creation of the decision boundary. For big datasets, the main emphasis should be given to the data processing, that is, cleaning and preprocessing are more relevant to extract useful information than adding more data. 
According to this, we seek the human intervention to be punctual and of consistent quality, that is, that it interacts with few data and, in our current implementation, it does so only with unstructured data. 


 
 
                                
 

  
 

 
 
 
 

As shown in Fig. \ref{fig:scanflow-extended-execution}, the Improver agent obtains the anomalous instances from the Checker, which are only a subset of the many instances that arrive in the inference stage. A human agent intervenes and labels a given amount of these instances (n\_labels). Then, the Improver applies geometric data augmentation (shear mapping), which is a non-intrusive technique that maintains the same semantics of the training distribution. The shear mapping is a linear transformation of $\mathbb{R}^{n}$ that distorts the shape but keeps the same n-dimensional measure (hypervolume) of any figure. For instance, in $\mathbb{R}^{2}$, the shear transformation applied to a rectangle becomes a parallelogram, preserving its area. Similarly, in $\mathbb{R}^{3}$, it preserves the volume. In particular, we applied this transformation to augment new instances that were previously labeled by a human, allowing the model to have more variety of examples to be trained, always keeping the same semantics.

At the end, the Improver applies a simple transfer learning from the current model's weights (new models fit using past weights) to retrain a new model after appending the new augmented data to the training set, thus emerging a more robust model in each human intervention. Transfer learning improves a target model on a target domain by transferring knowledge from a different source domain in order to reduce the dependence on large amounts of target domain data \cite{transferlearning}. For instance, a person that can drive a car can learn to drive a truck much faster than others, because both vehicles share some common knowledge. Similarly, the Improver takes the source domain knowledge learned from the training dataset and then transfers it to the target domain knowledge (dataset with noise) for constructing the improved model.



\subsubsection{Planner (Modify the current model in production)}
This agent is in charge of getting information from the Improver agent so that it can apply the required actions to the model (or workflow) if some conditions are met, for instance, if the model trustworthiness is greater than the previous one. The difference between this agent and the others, is that this one modifies the actual workflow, for example, it can replace the container that serves the model with another one. This procedure is described in Fig. \ref{fig:scanflow-extended-execution}. 


  

  
    
  
  

\section{Experiments and Results}
\label{sec:experiments-results}
\begin{figure}[htbp]
        \centering
    \includegraphics[width=0.95\linewidth]{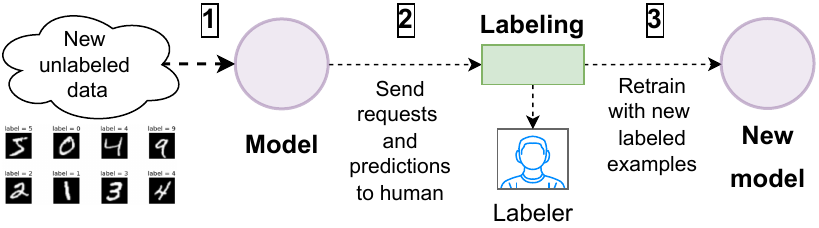}
    \caption{Pipeline for Scenario 2 (Random approach).}
    \label{fig:scenario2} 
\end{figure}

This section evaluates the effectiveness of using both human and machine agents to improve the model trustworthiness by comparing two scenarios. Scenario 1 ('Agents') corresponds to our proposal to retrain the model using information about data considered anomalous coming from Supervisor, Checker, and Improver agents (the Planner agent is meant for production systems) as described in Fig. \ref{fig:scenario1}. Scenario 2 ('Random'), which is fairly comparable to the first one, represents a common form of retraining a model. Instead of using the agents to get the anomalous points and augment them, it only performs labeling of random samples. That is, performs a random choice of the new instances, asks the human to label them, and retrains with (X\_train + X\_random, y\_train + y\_labels\_random). Fig. \ref{fig:scenario2} shows the pipeline for this scenario.

The methodology to compare these scenarios is as follows.
    First, we select a testing set from the original dataset and an inference set with noise for both scenarios. The model is evaluated on these two sets in order to make accuracy and NetTrustScore comparable. We also set the number of instances to be labeled (only 15 instances, even though the number of anomalous instances can be larger) and the seed number to have a fair comparison. 
    Second, for each iteration (time step), we issue a batch of new requests from the inference set to feed the model and evaluate how scenarios 1 and 2 perform on these new data. We repeat for a number of iterations while tracking the model improvement.


The testbed used in the experiments is as follows.
    Platform: Fedora Linux 35 (64 bits).
    Hardware: AMD Ryzen 9 5900HS, NVIDIA RTX 3060, 40 GB RAM.
    Software: Anaconda3 Python 3.8\footnote{https://www.anaconda.com} (Python data science platform), Tensorflow 2.5.0\footnote{https://www.tensorflow.org} (ML library), and Pytorch 1.9.0\footnote{https://www.pytorch.org} (ML library).
    Datasets: MNIST \cite{mnist} (60,000 28×28 pixel grayscale images of handwritten digits from 0 to 9), MNIST-C \cite{corruption} (corrupted MNIST with 4 corruptions: 'contrast', 'impulse', 'shot', and 'gaussian'), FashionMNIST \cite{fashionmnist} (60,000 28x28 pixel grayscale images of 10 types of clothing), and FashionMNIST-C \cite{corruption} (corrupted FashionMNIST with the same 4 corruptions). These datasets are considered relevant in image classification \cite{why-mnist1}.  

\begin{figure}[htbp]
\centerline{\includegraphics[width=0.85\linewidth]{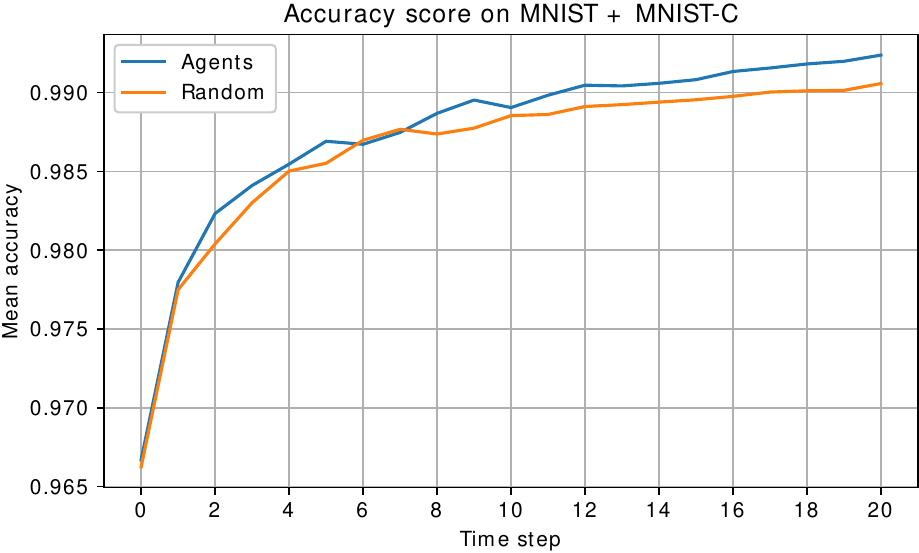}}
\caption{Accuracy evaluation on MNIST+MNIST-C dataset for 'Agents' and 'Random' scenarios.}
\label{fig:mnist_accuracy_avg}
\end{figure}

\begin{figure}[htbp]
\centerline{\includegraphics[width=0.85\linewidth]{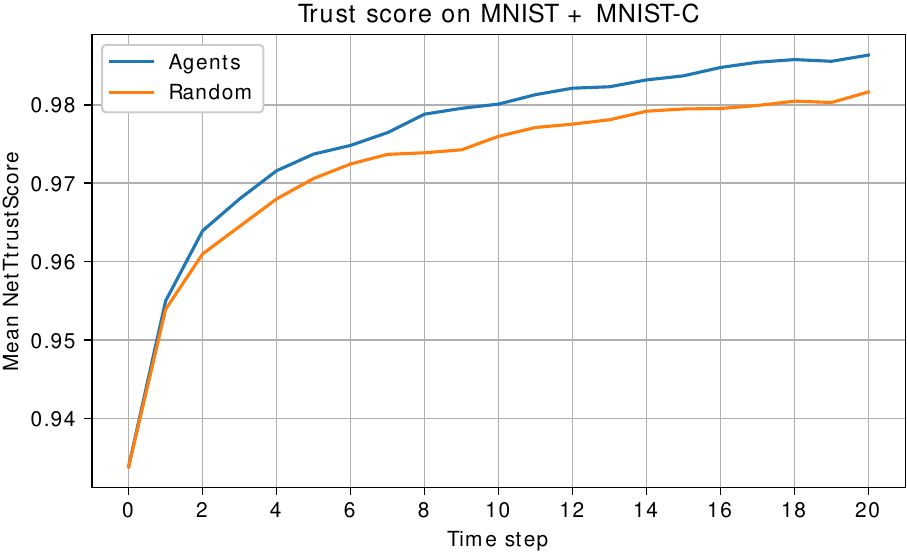}}
\caption{Trust evaluation on MNIST+MNIST-C dataset for 'Agents' and 'Random' scenarios.}
\label{fig:mnist_trust_avg}
\end{figure}

\begin{figure}[htbp]
\centerline{\includegraphics[width=0.85\linewidth]{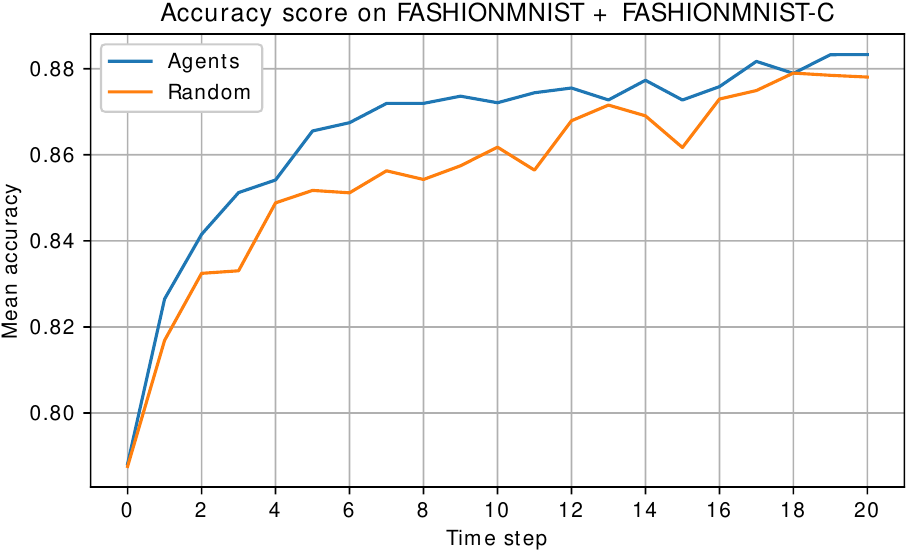}}
\caption{Accuracy evaluation on FashionMNIST+FashionMNIST-C dataset for 'Agents' and 'Random' scenarios.}
\label{fig:fashionmnist_accuracy_avg}
\end{figure}

\begin{figure}[htbp]
\centerline{\includegraphics[width=0.85\linewidth]{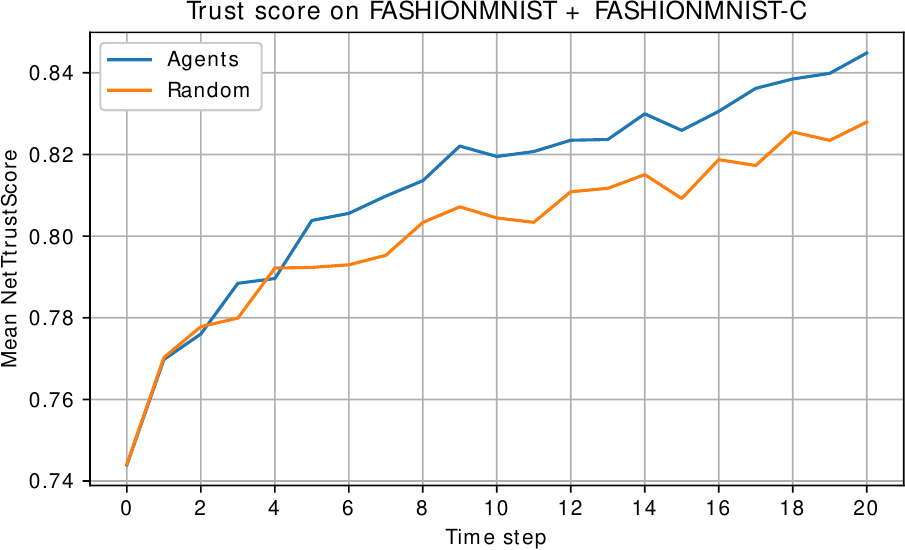}}
\caption{Trust evaluation on FashionMNIST+FashionMNIST-C dataset for 'Agents' and 'Random' scenarios.}
\label{fig:fashionmnist_trust_avg}
\end{figure}

As for the results on the MNIST dataset, Fig. \ref{fig:mnist_accuracy_avg} shows the accuracy on the two scenarios over 20 iterations. We can see that the model accuracy improves on both scenarios over time, being the 'Agents' approach the one that performed better. Similarly, as shown in Fig. \ref{fig:mnist_trust_avg}, which compares the trust on both scenarios, the 'Agents' approach outperforms the 'Random' method, basically because the set of instances used to retrain the model (anomalous instances) were more significant in this scenario, hence, the model generalized better. Apart from that, the joint work to detect anomalous instances, augment data, and transfer the knowledge are the key contributions from the agents. Although they do not learn its abilities by themselves, the communication and custom-made skills as a whole bring an interactive approach to make the agents interact with its environment (human, model, metadata, etc.) to improve the model trust. 
Moreover, the 'Agents' approach also shows better results on the FashionMNIST dataset. As shown in Fig. \ref{fig:fashionmnist_accuracy_avg}, the accuracy also gets improved on each iteration. Similarly, as shown in Fig. \ref{fig:fashionmnist_trust_avg}, the trust score also increases more than in the 'Random' approach.

\begin{figure}[htbp]
\centerline{\includegraphics[width=0.85\linewidth]{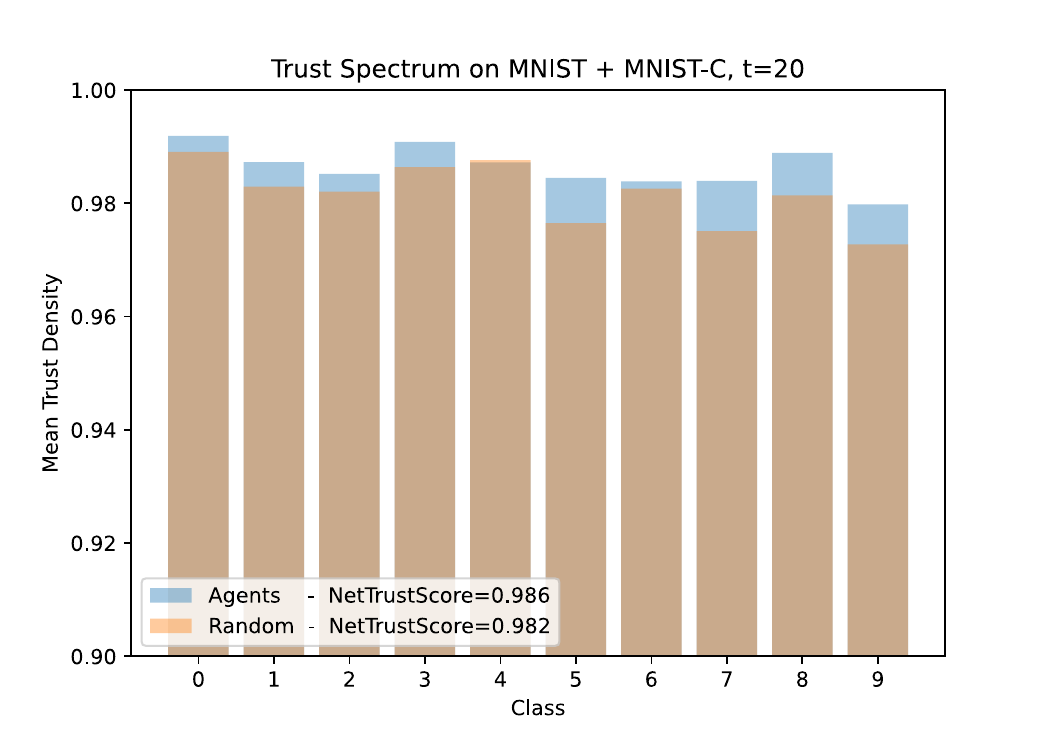}}
\caption{Trust Spectrum on MNIST+MNIST-C dataset for 'Agents' and 'Random' scenarios for the latest model (timestep=20).}
\label{fig:trust_sp_1}
\end{figure}

\begin{figure}[htbp]
\centerline{\includegraphics[width=0.85\linewidth]{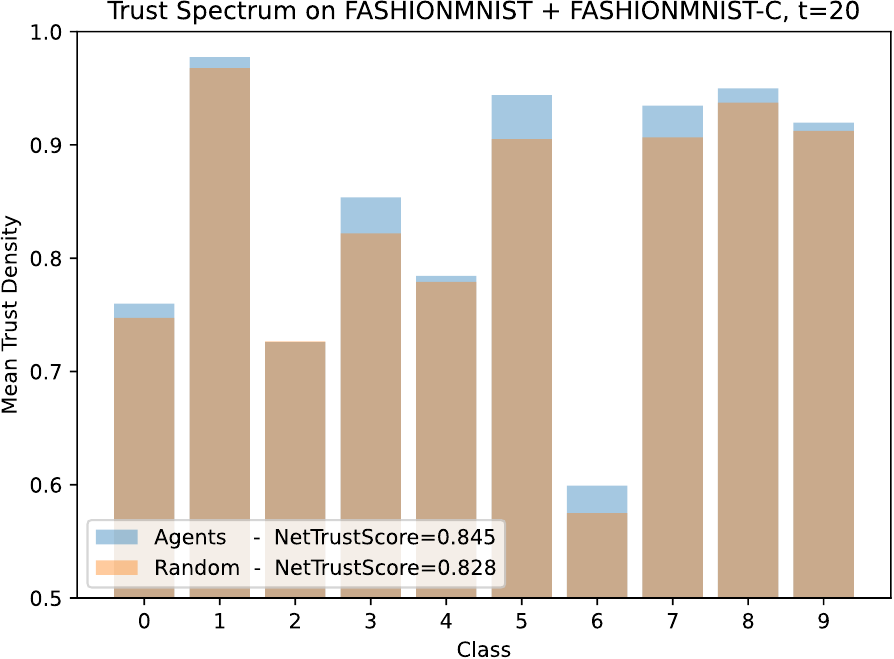}}
\caption{Trust Spectrum on FashionMNIST+FashionMNIST-C dataset  for 'Agents' and 'Random' scenarios for the latest model (timestep=20).}
\label{fig:trust_sp_2}
\end{figure}

It is also relevant to analyze the improvement in disaggregated form, that is, how each new model behaves in each class (label). Figs. \ref{fig:trust_sp_1} and \ref{fig:trust_sp_2} show the trust spectrum for the 'Agents' approach (blue area) and the method using random choice (brown area). Regarding MNIST dataset, Fig. \ref{fig:trust_sp_1} shows that the model trust improves slightly in almost all the classes. Although on average the increase in trust is not that much, it indicates that certain new instances were more significant (anomalous instances) and the model can continuously improve to reduce its overconfidence.

Regarding FashionMNIST dataset, Fig. \ref{fig:trust_sp_2} shows that the improvement is greater in general, and especially in class 3 and class 5. So, we can say that not only the 'Agents' approach provides a higher number of correct answers (accuracy) but also is very confident about these answers. Due to the higher complexity of the dataset, the trust could be improved considerably, indicating that it is not the quantity of instances that matters but the quality of the instances to be labeled. However, we observe that both methods have low trust in class 6, indicating that further research should be done on this point for the production deployment of such models.

Finally, as for the execution time, the procedure triggered on each iteration for scenarios 1 and 2 took about 20 seconds and 6s in average, respectively, for both MNIST and FashionMNIST. Since the procedure is executed in the background, the difference is not very relevant.

\section{Conclusions and Future work}
\label{sec:conlusions-future-work}
In this paper, we have discussed how to iteratively improve trust in a ML model. For this purpose, we proposed a multi-agent system that includes human and machine agents that interact with each other to detect anomalous instances through a symbolic rule based on the trust score of individual instances, augments them, and performs a retraining transferring the knowledge of the previous model. We evaluated this system using two well-known datasets in the field of computer vision, but our approach can be extrapolated to other datasets and ML fields since there is no modification in the model structure.

We showed that the great advantage of our agents is in their joint work on the model. The feedback that flows in each agent is the fundamental piece that improves both accuracy and trust. In contrast, the improvement in the random approach comes only from the labels that were added to the retraining. 

Clearly, continuous intervention in ML models allows improving their behavior fast and efficiently. On the one hand, monitoring the trust score of each instance during the inferences allows detecting any drift in the confidence of the predictions. On the other hand, well-defined rules contribute to guarantee the behavior of the model, in this case, by identifying the critical instances to be considered unsafe. Additionally, expanding the data allows dealing with few information to update the model and also helps to improve its generalization. Whereas data augmentation could be more complex, we showed that it works to improve trust.

As future work, we will consider use cases beyond classification, such as regression, clustering, and association rule learning, among others. We also plan to create a way of learning agent behaviors. That is, start from a knowledge base in each agent (for example, the Checker would start with a certain ability to detect anomalies), and enable them to enhance their behavior to increase the total reward using individual improvements over time. This scheme would follow a reinforcement learning approach, where actions are learned using information from the interaction with the environment. However, due to the limited feedback that may be available when putting a model into production, an alternative might be using data in offline mode, i.e., interacting with logged experience (also called, offline reinforcement learning). This new feature can evolve the multi-agent system to adapt to new circumstances that the model may undergo.

\section*{Acknowledgment}
This work was supported by Lenovo as part of Lenovo-BSC 2020 collaboration agreement, by the Spanish Government under contracts PID2019-107255GB-C21 and PID2019-107255GB-C22, and by the Generalitat de Catalunya under contract 2017-SGR-1414 and under grant 2020 FI-B 00257.

\bibliographystyle{IEEEtran}
\bibliography{IEEEabrv,biblio}


\end{document}